\documentclass[10pt,letterpaper]{article}
\usepackage[top=0.85in,left=2.75in,footskip=0.75in]{geometry}

\usepackage{amsmath,amssymb}

\usepackage{changepage}

\usepackage{textcomp,marvosym}

\usepackage{cite}

\usepackage{nameref,hyperref}

\usepackage[right]{lineno}

\usepackage[nopatch=eqnum]{microtype}
\DisableLigatures[f]{encoding = *, family = * }

\usepackage[table]{xcolor}

\usepackage{array}

\usepackage{makecell}
\usepackage{multirow}

\newcolumntype{+}{!{\vrule width 2pt}}

\newlength\savedwidth

\raggedright
\usepackage[aboveskip=1pt,labelfont=bf,labelsep=period,justification=raggedright,singlelinecheck=off]{caption}

\makeatletter
\renewcommand{\@biblabel}[1]{\quad#1.}
\makeatother

\usepackage{lastpage,fancyhdr,graphicx}
\usepackage{epstopdf}
\fancyheadoffset[L]{2.25in}
\fancyfootoffset[L]{2.25in}
\begin{document}
\vspace*{0.2in}

\begin{flushleft}
{\Large
\textbf\newline{Self-Contrastive Weakly Supervised Learning Framework for Prognostic Prediction Using Whole Slide Images} 
}
\newline
\\
Saul Fuster\textsuperscript{1\Yinyang*},
Farbod Khoraminia\textsuperscript{2\Yinyang},
Julio Silva-Rodríguez\textsuperscript{3},
Umay Kiraz\textsuperscript{4,5},
Geert J. L. H. van Leenders\textsuperscript{6},
Trygve Eftestøl\textsuperscript{1},
Valery Naranjo\textsuperscript{7},
Emiel A. M. Janssen\textsuperscript{4,5},
Tahlita C. M. Zuiverloon\textsuperscript{2,},
Kjersti Engan\textsuperscript{1}
\\
\bigskip
\textbf{1} Dept. of Electrical Engineering and Computer Science, University of Stavanger, 4021 Stavanger, Norway
\\
\textbf{2} Dept. of Urology, Erasmus MC Cancer Institute, University Medical Center, 3015 GD Rotterdam, The Netherlands
\\
\textbf{3} ÉTS Montréal, Montréal, Québec 1011, Canada
\\
\textbf{4} Dept. of Pathology, Stavanger University Hospital, 4011 Stavanger, Norway
\\
\textbf{5} Dept. of Chemistry, Bioscience and Environmental Engineering, University of Stavanger, 4021 Stavanger, Norway
\\
\textbf{6} Dept. of Pathology and Clinical Bioinformatics, Erasmus MC Cancer Institute, University Medical Center Rotterdam, 3015 GD Rotterdam, The Netherlands
\\
\textbf{7} CVBLab, Instituto Universitario de Investigación en Tecnología Centrada en el Ser Humano (HUMAN-tech), Universitat Politècnica de València, 46022, Valencia, Spain
\\
\bigskip

%
%
\Yinyang These authors contributed equally to this work.





* saul.fusternavarro@uis.no

\end{flushleft}
\section*{Abstract}
We present a pioneering investigation into the application of deep learning techniques to analyze histopathological images for addressing the substantial challenge of automated prognostic prediction. Prognostic prediction poses a unique challenge as the ground truth labels are inherently weak, and the model must anticipate future events that are not directly observable in the image. To address this challenge, we propose a novel three-part framework comprising of a convolutional network based tissue segmentation algorithm for region of interest delineation, a contrastive learning module for feature extraction, and a nested multiple instance learning classification module. Our study explores the significance of various regions of interest within the histopathological slides and exploits diverse learning scenarios. The pipeline is initially validated on artificially generated data and a simpler diagnostic task. Transitioning to prognostic prediction, tasks become more challenging. Employing bladder cancer as use case, our best models yield an AUC of 0.721 and 0.678 for recurrence and treatment outcome prediction respectively.

\section*{Author Summary}
Predicting disease outcomes, or prognostic prediction, is more challenging than diagnosis due to factors like tissue variability and treatment response diversity. Urinary bladder cancer prognosis, in particular, is complicated by heterogeneous behaviors and lacking clear markers. While diagnostic research is prevalent in computational pathology, prognostic prediction receives less attention due to data complexities. Our study introduces an automated method leveraging deep learning to predict urinary bladder cancer prognosis on large unlabelled data. By focusing on key tissue areas and utilizing advanced machine learning and image processing techniques, our approach enhances prognostic accuracy. This work marks a significant advancement in urinary bladder cancer prognosis, laying the foundation for future research.

\section*{Introduction}
The introduction of digital pathology, characterized by the digitization of tissue sections into whole slide images (WSI) through microscopy scanners, has opened up numerous possibilities. A prevailing trend in computational pathology (CPATH) research involves utilizing image processing and machine learning to develop tools that assist pathologists in visualization, region of interest (ROI) extraction, and diagnostic tasks \cite{van2021deep}.

Prognostic prediction is generally acknowledged as more challenging than diagnostic prediction as it involves forecasting future events and outcomes. Prognosis is arduous due to histological diversity, observer variability, and tumor heterogeneity \cite{madabhushi2016image}. Additionally, diverse treatment responses and recurrence patterns must be considered. Urinary bladder prognostic prediction exemplifies this complexity as a result of diverse treatment responses, recurrence patterns, and the absence of distinct markers and varied clinical factors \cite{babjuk2022european}.

Plenty of research in CPATH is dedicated to diagnostic prediction, often relying on manually selected ROIs during both model learning and inference stages. For prognostics, weakly labeled data is commonly employed, with patient-based labels defining treatment outcomes, recurrence, or disease progression. Consequently, there is no guarantee that a particular region contains the necessary information, or that the essential information is genuinely present in the WSI.

In this study, we present an automated deep learning pipeline for prognostic predictions from WSI, trained and tested on weakly labeled data. A nested multiple instance learning method with attention mechanisms (NMIA), recently proposed by our research group, is used for the first time in an end-to-end problem \cite{fuster2022nested}. The pipeline utilizes extensive unannotated regions to enhance feature representations and explores the impact of selectively choosing informative areas within the slides. This work represents a pioneering investigation into the application of prognostic predictions in urinary bladder cancer. The main contributions of this paper are summarized as:

\begin{enumerate}
    \item Introduction of an automated deep learning pipeline for prognostic predictions from WSI, leveraging weakly labeled data. The pipeline incorporates NMIA, a novel approach applied for the first time in an end-to-end problem.
    \item Utilization of extensive unannotated regions to enrich feature representations and investigation into the impact of selectively choosing informative areas within the slides guided by domain knowledge, marking a pioneering exploration into the application of prognostic predictions in urinary bladder cancer.
\end{enumerate}

\section*{Background \& Related Work}
\subsection*{Urinary Bladder Cancer}

Non-muscle invasive bladder cancer (NMIBC) conform approximately 75\% of bladder cancer diagnoses \cite{burger2013epidemiology}. The 2022 version of the European Association of Urology (EAU) guidelines on NMIBC suggests that patients are stratified into risk groups based on the hazard to progress to a muscle-invasive disease \cite{babjuk2022european}. The hazard score depends on significant clinical and pathological factors, which are themselves time-consuming to determine and frequently result in variations among uropathologists. Currently, intravesical Bacillus Calmette–Guérin (BCG) is the gold standard adjuvant therapy for high-risk non-muscle invasive bladder cancer (HR-NMIBC). Unfortunately, BCG treatment causes many severe adverse reactions, and up to 50\% of the patients will develop BCG resistance, thus resulting in recurrence or progression \cite{kamat2016definitions}. Identifying these patients at the first stage would significantly reduce the recurrence rate and treatment cost, hence contributing to more adequate patient-based treatment strategies. Nonetheless, bladder WSI are sizable and often contain disorganized, fragmented tissue sections, with a significant number of artifacts and other non-diagnostically-relevant tissue \cite{burger2013epidemiology}. The unique challenges of urinary bladder cancer WSI have served as inspiration for the three-step pipeline presented in this study. Also, to the best of our knowledge, a deep learning system for BCG response prediction relying on image features does not exist in the literature.

\subsection*{Data Modalities}
Computer-aided diagnosis (CAD) systems that utilize machine learning techniques for medical imaging analysis of WSI have shown effective ways to reduce subjectivity and speed up the diagnostic process \cite{van2021deep}. WSI are pre-stored at various magnification levels, allowing pathologists to quickly adjust the zoom level, analogous to physical microscopes. Lower magnification is typically used to view tissue-level morphology, while higher magnification is useful for examining cell-level features. In CPATH, imaging techniques that rely on convolutional neural networks (CNNs) are considered the best option for extracting features from histological images \cite{cui2021artificial}. However, prognostic applications have primarily relied on clinicopathological information more than on image data \cite{kourou2015machine}. Both clinicopathological information and image features are derived from the visual characteristics of tumor tissue. Image features are extracted directly from the image, while clinicopathological information depends on external observations. Nonetheless, recent deep learning prognostic applications employ image features, while providing reasonably accurate prognostic predictions. Although there are methods based on clinicopathological or image data alone, hybrid solutions have also been proposed with state-of-the-art performance \cite{zheng2022accurate, tokuyama2022prediction}. Another promising type of data is that of genomics, specifically next-generation sequencing, which has altered the understanding and assessment of cancer\cite{de2023non}. However, next-generation sequencing is an expensive technology, still in its early implementation phase, making it an inaccessible solution for many pathology laboratories at present. In contrast to this, hematoxylin and eosin-stained (H\&E) WSI provide an affordable solution and practical choice for routine diagnostic and research purposes. Therefore, we explore H\&E WSI both with and without clinical information in this study.

\subsection*{Non-Supervised Learning Methods}
One challenge in CPATH is the lack of WSI with detailed or region of interest (ROI) annotations, which is often expensive and time-consuming to acquire. To address this, weakly supervised methods have been proposed for training deep learning models on WSI \cite{van2021deep, wang2019weakly, zhou2021histopathology, wang2022weakly, andreassen2023deep}. Weakly supervised methods emerge as an advantageous approach for prognostic applications as they accommodate the uncertainty and heterogeneity of medical data. Among the diverse weakly supervised methods, attention-based multiple instance learning (AbMIL) is a popular approach \cite{ilse2018attention, lu2021data, srinidhi2021deep, zhang2022dtfd}. The method uses an attention mechanism to selectively focus on regions of interest within an image, allowing the model to learn from weakly labeled data. One diagnostic application using weakly supervised deep learning on WSI is that of predicting the pathological grade of the patient \cite{campanella2019clinical, zhang2019pathologist, wetteland2021automatic}. Such approach was able to achieve performance on par with supervised methods, while reducing the amount of annotated data required. Cancer survival prediction using WSI was performed with AbMIL, as demonstrated in \cite{yao2020whole, le2023deep, liu2024advmil, liu2023dsca, godson2024immune}. It is highlighted that the attention-based approach improved the performance of the model. However, WSI present scattered tissue, with countless instances. Hence, a straightforward MIL approach may not be suitable, as WSI could be densely populated with noisy instances. This led us to propose NMIA, which restricts cross-contamination among regions within the images \cite{fuster2022nested}.

The relationship between image features and patient outcomes can be complex and difficult to discern. To address this lack of correlation, self-supervised methods can overcome the disparity. In recent years, contrastive learning has emerged as a promising technique for learning feature representation from large unlabeled datasets. Contrastive learning is a type of learning with the aim of training a feature extractor, using a contrastive loss function \cite{chen2020simple}. This learning approach has the capacity to utilize extensive unannotated regions for enhancing feature representation. Typically, a contrastive module serves as a preliminary step before classification, facilitating the extraction of feature representations \cite{ciga2022self, fashi2022self}. Moreover, the acquisition of transformation-independent features has proven effective in mitigating stain variation \cite{ke2021contrastive, perez2022staincut}. It has also been employed for maximizing feature similarity between areas from the same WSI \cite{li2021dual, li2022lesion}. Contrastive learning in WSI prognostics is mostly unexplored \cite{tu2022dual}. We adopt SimCLR and variations for exploiting underlying prognostic patterns in WSI \cite{chen2020simple}.

\section*{Dataset}
In this study, we have gathered NMIBC WSI from two distinct cohorts. The total number of patients per application is displayed in Table \ref{tab:patients}.

\begin{table}[h]
\caption{\textbf{Description of the patient sets, $S_{\text{EMC}}$ and $S_{\text{SUH}}$.}}
\centering
\begin{tabular}{|c|cc|cc|}
\hline
\multirow{2}{*}{\textbf{Set}}    & \multicolumn{2}{c|}{$S_{\text{EMC}}$} & \multicolumn{2}{c|}{$S_{\text{SUH}}$} \\ \cline{2-5} 
                                 & BCG-R     & BCG-NR     & Rec                 & NoRec                 \\ \hline \hline
\multicolumn{1}{|c|}{Train}      &    272 (72)           &   81 (42)                &      113 (0)               &        107 (0)            \\ \cline{1-1}
\multicolumn{1}{|c|}{Validation} &    25 (24)           &     25 (22)              &        18 (0)             &         12 (0)           \\ \cline{1-1}
\multicolumn{1}{|c|}{Test}       &     25 (25)          &     25 (25)              &          27 (0)           &           23 (0)         \\ \hline
\end{tabular}
\caption*{\textit{The number between parenthesis indicates the number of patients with annotated WSI.}}
\label{tab:patients}
\end{table}

We have available HR-NMIBC WSI from a multi-centre cohort provided by Erasmus Medical Center (EMC), Rotterdam, The Netherlands. We denote this dataset $S_{\text{EMC}}$, and use it for BCG response prediction. Let BCG-R and BCG-NR denote BCG responder and non-responder tumor, correspondingly. BCG-NR corresponds to BCG failure according to EAU guidelines, excluding BCG intolerance. A total of 453 patients and 503 WSI formed the dataset. Since the treatment outcome is related to the patient as a whole and not to a specific region of the tumor, patches from various WSI that belong to the same patient were merged as a single entity. Not all slides contained annotations, and among those that were annotated, none of the WSI were fully annotated due to time and database storage constraints. Annotations contain tissue types, artifacts, grading and staging. Moreover, a detailed report of clinicopathological information per patient was disclosed with information about BCG treatment guidelines, pathological diagnoses, and patient demographics. We utilized clinical variables of gender, age, smoking status, grade, stage, concomitant carcinoma in situ, size and focality of the tumor.

We also included a total of 300 WSI corresponding to 300 different NMIBC patients from Stavanger University Hospital (SUH), Stavanger, Norway. We denote this dataset $S_{\text{SUH}}$, and use it for recurrence prediction. Let NoRec and Rec denote no recurrence and recurrence, respectively. Rec was defined as recurrent tumors in the bladder only, with a median follow-up of 82 months. No WSI were annotated with ROI, but weak labels regarding recurrence outcome were available.

\begin{figure*}[h]
\centerline{\includegraphics[width=\columnwidth]{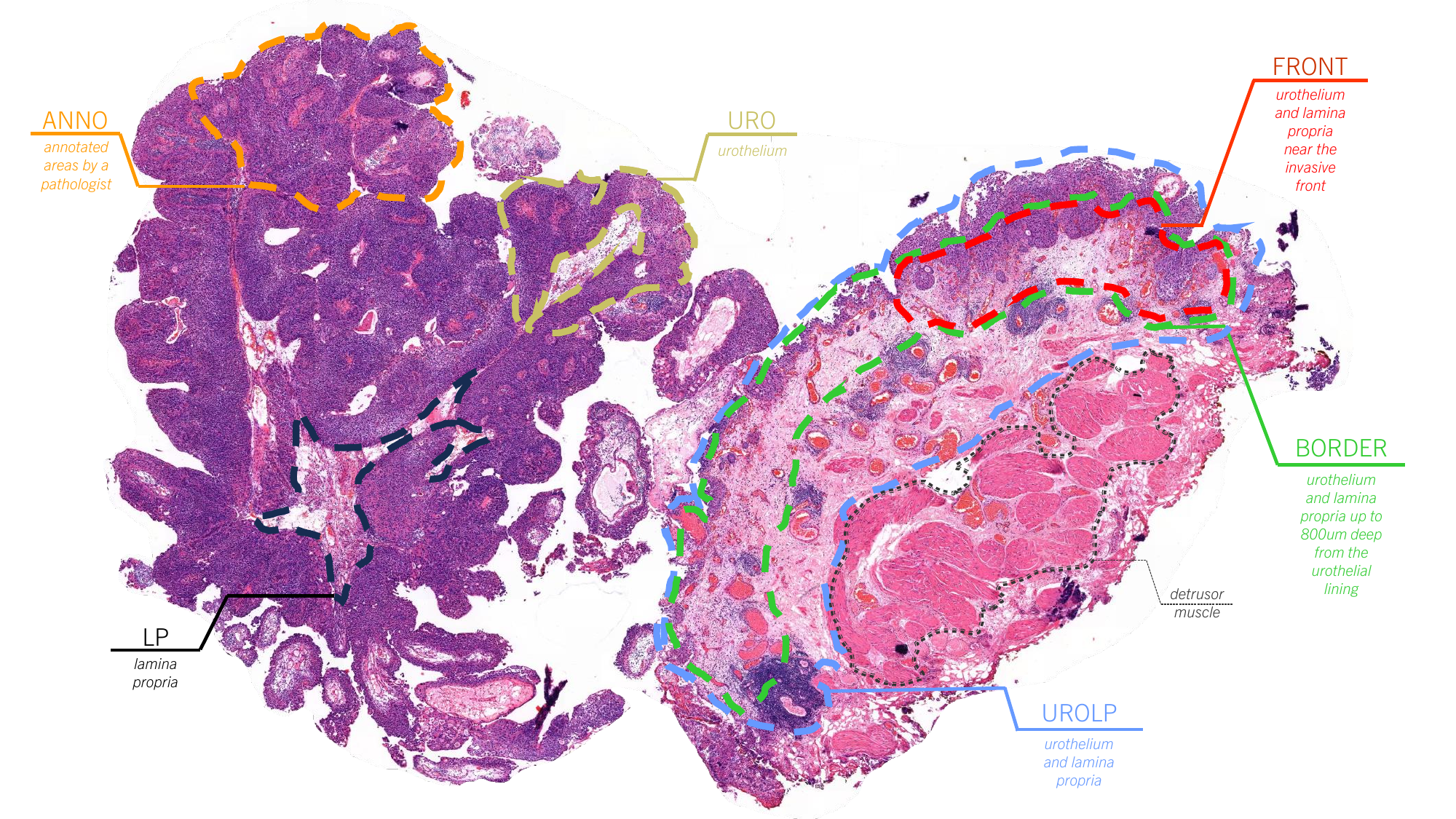}}
\caption{\textbf{Overview of ROIs utilized.} The process of extracting ROIs from raw WSI involved either a tissue segmentation algorithm and/or pathologist's annotations. The annotations highlighted areas that were deemed prognostically significant for predicting outcome, while the algorithm provided masks that highlighted different tissue types. Subsets of tissue were extracted using urothelium and lamina propria. For magnification levels, the study explored two mono-scale approaches using 10x and 20x, as well as a multi-scale method using three magnifications (2.5x, 10x, 40x).}
\label{fig:datasets}
\end{figure*}

With regards to ROI definition, two main strategies were employed: using annotated areas or areas defined from an automatic tissue segmentation algorithm. Fig. \ref{fig:datasets} highlights the various ROIs explored in this study. The automatic definition of ROIs is later described in \nameref{sec:segmentation}. Concerning annotations, an engineer with specific training in bladder pathology (F. K.) partially annotated 217 of the total 503 EMC patient slides from $S_{\text{EMC}}$, under the supervision of an experienced uropathologist (G. vL.). This subset of data is referred as $D_{\text{ANNO}}$. The annotation process consisted of general tissue type annotations, and in some cases, sub classes indicating grading, presence of tumor infiltrating lymphocytes, flat lesions, and invasive areas. First, a coarse annotation of the WSI was done, identifying and labeling the main regions of interest within the images. Then, a quality control process was implemented to ensure the annotation's quality and consistency by getting external revision from expert uropathologists.

\subsection*{Compliance with Ethical Standards}
This work involved human subjects in its research. Approval of all ethical and experimental procedures and protocols was granted by the Regional Committees for Medical and Health Research Ethics (REC), Norway, ref.no.: 2011/1539, regulated in accordance to the Norwegian Health Research Act; and the Daily Board of the Medical Ethics Committee Erasmus MC, Rotterdam, The Netherlands, METC number: MEC-2019-704.


\section*{Methods}
\begin{figure*}[h]
\centerline{\includegraphics[width=\columnwidth, trim={0.25cm 3cm 0.25cm 2.5cm},clip]{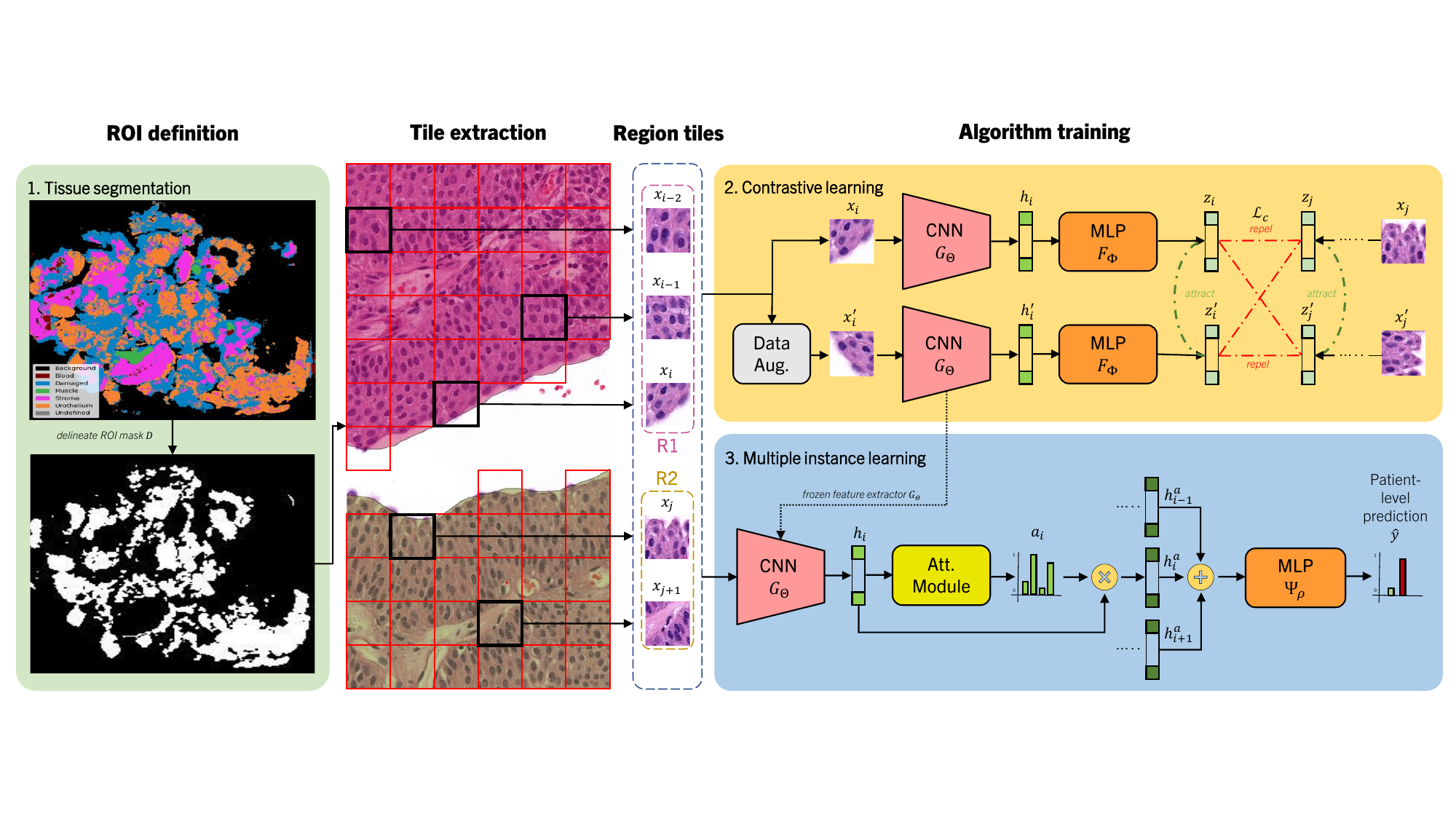}}
\caption{\textbf{Deep learning pipeline for prognostic outcome prediction.} 1) A tissue segmentation is employed for delineating a ROI of choice $D$. Then,tiles are extracted from WSI regions for training an algorithm. 2) Contrastive learning is employed to learn representations of the tiles. 3) The representations are then used to train an AbMIL model that predicts the prognostic outcome. This approach compresses an end-to-end pipeline where the raw input image data is broken down and processed for predicting clinical outcome.}
\label{fig:pipeline}
\end{figure*}

In the upcoming subsections, we introduce a three-step fully-automated pipeline for WSI prognosis that combines region of interest (ROI) extraction, contrastive learning for feature representation and multiple instance learning (MIL) for predicting the concluding prognostic outcome, as depicted in Fig \ref{fig:pipeline}. This approach enables us to optimize the model by leveraging the benefits of these techniques. It ensures the inclusion of important instances for predicting clinical outcome from WSI visual cues, while maintaining computational feasibility. Ultimately, the proposed steps for prognostic predictions in histopathological imaging are the following:

\begin{enumerate}
    \item Define and extract ROIs using a tissue segmentation algorithm for tile extraction strategies.
    \item Train a feature extractor $G_\theta$ to generate an intermediate dataset $\mathcal{H}$ using contrastive learning.
    \item Use image feature embeddings $\mathcal{H}$ for prognostic classification using MIL.
\end{enumerate}

\subsection*{Automatic Region of Interest Segmentation}
\label{sec:segmentation}
\begin{figure*}[h]
\centerline{\includegraphics[width=\columnwidth, trim={2.25cm 4.25cm 2.25cm 4.45cm},clip]{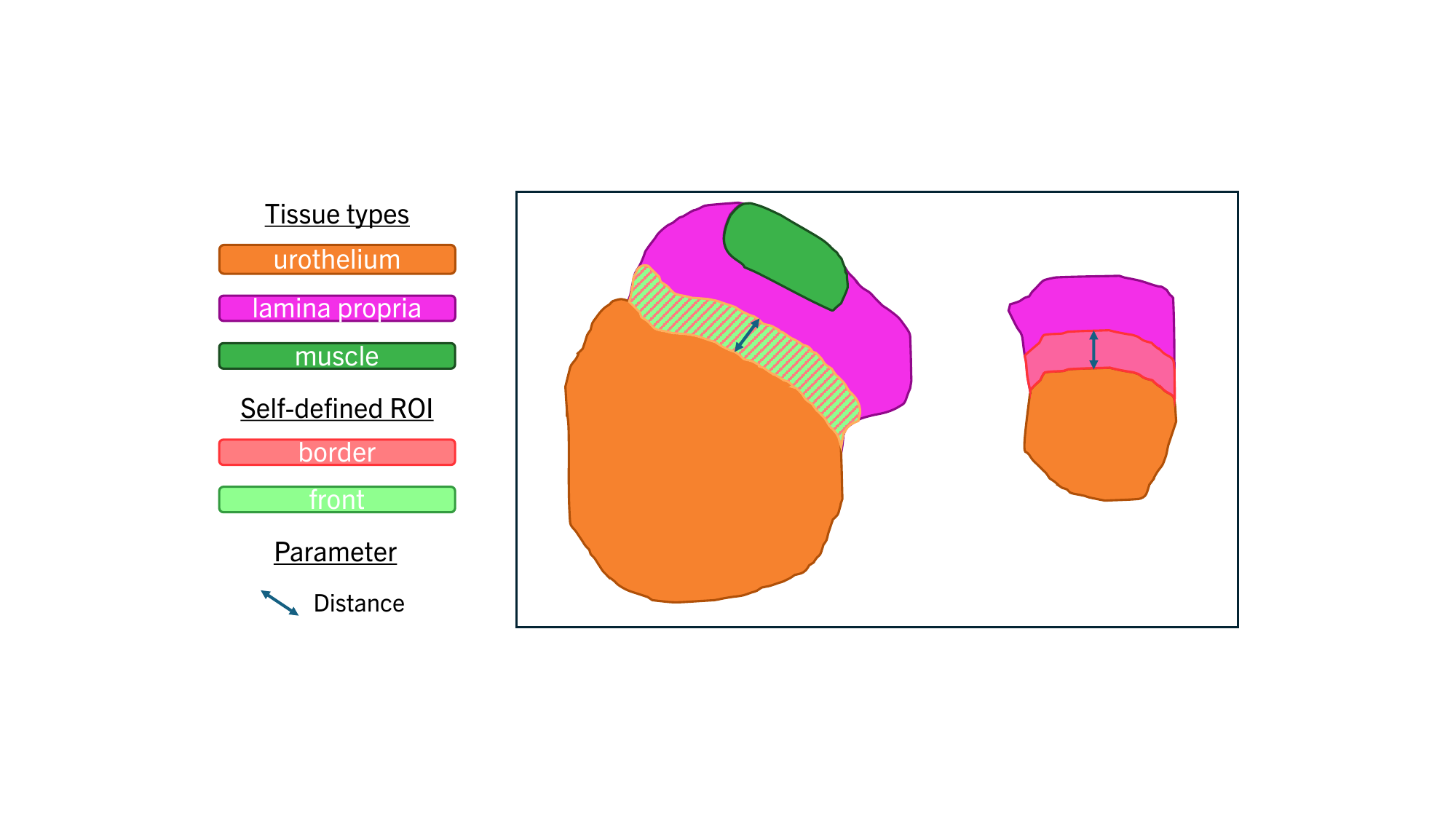}}
\caption{\textbf{Schematic representation of self-defined ROI generation.} Guided by the segmentation mask of various tissue types, we apply diverse image processing morphological operations to define ROIs $D_{y}$ based on domain knowledge from expert pathologists. In the example displayed, we enlarge the urothelium and lamina propria masks applying dilation, limited by a distance parameter determined on clinical expertise. The resulting overlapping areas represent $D_{\text{BORDER}}$. We further delineate a subset of $D_{\text{BORDER}}$ by extracting only those areas where muscle is present within the same tissue section, aiming to represent the potential invasive front in $D_{\text{FRONT}}$.}
\label{fig:schematic}
\end{figure*}

The current study will explore various ROI configurations as the localization of the tissue of interest is unknown. While annotations can be expensive and inflexible, automatic tissue segmentation algorithms provide the flexibility to redefine ROIs based on different clinical considerations. A tissue segmentation algorithm was proposed in \cite{wetteland2020multiscale} for $S_{\text{SUH}}$, while an active learning-based approach for tissue segmentation was developed in \cite{fuster2023active} for $S_{\text{EMC}}$. Both models share the same architecture, utilizing a tri-scale CNN backbone that leverages different magnifications for each input CNN. The tissue segmentation algorithms work at patch level, and classify all patches $x$ in the WSI as $y \in {\cal{Y}}=$ \{urothelium, lamina propria, muscle, blood, damage, background\}. The dataset resulting from extracting patches with label $y$ is denoted $D_{y}$.  For example, urothelium tissue is the most prominent source of information in urothelial bladder carcinoma, and the dataset of patches extracted from these regions are denoted $D_{\text{URO}}$. In conjunction with urothelium, lamina propria may serve an important role for influencing the growth of the tumor \cite{andersson2014lamina}. The dataset of patches extracted from lamina propria is denoted $D_{\text{LP}}$, and the union urothelium and lamina propria $D_{\text{UROLP}} = D_{\text{URO}} \cup D_{\text{LP}}$. However, these $D_{y}$ are solely defined based on tissue types, and to meticulously analyze and concentrate on the pertinent area of interest for comprehending the disease's status, it is imperative to define tailored ROIs. Consequently, exploiting domain knowledge through segmentation maps is a pivotal aspect of this work. Notably, not all lamina propria might be interacting with the tumor. Therefore, we defined a depth of 800$\mu m$ based on medical knowledge for defining possible tumor and immune response interactions based on medical knowledge. The union of the boundary between urothelium and lamina propria defines the dataset $D_{\text{BORDER}} \subset D_{\text{UROLP}}$. This is accomplished by applying a disk dilation operation on the urothelium and lamina propria masks, where the disk radius is determined by pixel size, and subsequently segment the overlapping area to extract the bordering region. Fig. \ref{fig:schematic} displays an schematic representation of the self-defined ROI $D_{\text{BORDER}}$. Furthermore, the invasive front of the tumor, which represents the most aggressive part of the tumor, could potentially provide the most significant features for comprehending the current state of the tumor in relation to the patient's immune system \cite{brooks2016positive}. Therefore, we refine $D_{\text{BORDER}}$ using a region-growing algorithm along these borders to exclude areas lacking muscle tissue within a tissue section, thus defining $D_{\text{FRONT}} \subset D_{\text{BORDER}}$. To exclude distant muscle areas from consideration, we apply the same distance threshold of 800$\mu m$, thus ensuring the focus remains on the invasive front regions.

Regarding tile extraction strategies, various magnification levels and tile sizes are used. In the case of mono-scale models, we used a tile size of $256\times256$ for 10x magnification and $512\times512$ for 20x magnification. This decision was made to ensure that the patches covered the same physical area, i.e. field of view. In the case of multi-scale models TRI, we used a tile size of $128\times128$ for all three magnification levels (40x, 10x, and 2.5x), extracted following the approach described in \cite{wetteland2021parameterized}.

\subsection*{Feature Extraction via Contrastive Learning}
\label{sec:feature}
Contrastive learning is a specific approach within metric learning that focuses on comparing similar and dissimilar pairs of examples, based on a siamese structure, by attracting and repelling representations accordingly \cite{chen2020simple}. A contrastive model is trained as follows: given a batch of $N$ random samples from a training set of $X$ images, a set of two image transformations are applied to all images in the batch in order to obtain $2N$ augmentations. These transformed images are forwarded though a feature extractor $G_\theta : {\mathcal{X}} \to \mathcal{H}$ and projected into a low-dimensional feature space through a multi-layer perceptron (MLP) $F_\phi : {\mathcal{H}} \to \mathcal{Z}$, to be later $\textit{l}_2$ normalized. Let $\mathrm{sim}(\mathbf{z}_i,\mathbf{z}_j)=\mathbf{z}_i^{\top} \mathbf{z}_j / \left\| \mathbf{z}_i \right\| \left\| \mathbf{z}_j \right\|$ be the cosine similarity between $\textit{l}_2$ normalized vectors $\mathbf{z}_i$ and $\mathbf{z}_j$, then the loss function for a positive pair is defined as:

\begin{equation}
\mathcal{L}_c  = -\frac{1}{2N}\sum_{i\in I}^{} \mathrm{log}\frac{\mathrm{exp}(\mathrm{sim(\mathbf{z}_i,\mathbf{z}_i{'})/ \tau)}}{\sum_{j=1}^{2N} 1_{[j\neq i]}\mathrm{exp}(\mathrm{sim(\mathbf{z}_i,\mathbf{z}_j)/ \tau)}}
\label{eq:cl}
\end{equation}

where $\mathbf{z}_i{'}$ is the augmented representation of $\mathbf{z}_i$, $1_{[j\neq i]}$ is a binary indicator to indicate all other instances than $i$, and $\tau$ is a temperature coefficient to control the strenght of penalties on hard negative samples. The contrastive loss has the objective of ensuring the model learns strong feature representations regardless of the augmentation applied, thus increasing the robustness to variability in the input image. The contrastive loss rewards the model for creating similar features for both augmentations from the same image, while increasing feature dissimilarity between other augmentations from images in the batch. $\mathcal{L}_c$ corresponds to the unsupervised version, although the supervised contrastive learning loss $\mathcal{L}_{sc}$ does exist \cite{khosla2020supervised}. The supervised loss $\mathcal{L}_{sc}$ considers images from the same class in the batch, using the corresponding image label $y$, and does not punish the model for generating similar representations among images from the same class:

\begin{equation}
\mathcal{L}_{sc}  = \sum_{i\in I}^{}\frac{-1}{|P_{(i)}|} \sum_{p\in P_{(i)}}^{} \mathrm{log}\frac{\mathrm{exp}(\mathrm{sim(\mathbf{z}_i,\mathbf{z}_i{'})/ \tau)}}{\sum_{j=1}^{2N} 1_{[j\neq i]}\mathrm{exp}(\mathrm{sim(\mathbf{z}_i,\mathbf{z}_j)/ \tau)}}
\label{eq:scl}
\end{equation}

where $P_{(i)} \equiv p : y_p = y_i$ , while $|P_{(i)}|$ represents its cardinality. $\mathcal{L}_{sc}$ is preferred when labeled data is available, allowing for explicit learning of relevant representations and improved model performance. However, when labeled data is scarce or difficult to obtain, $\mathcal{L}_{c}$ can be a practical option.

We further explore the implementation of contrastive learning defining a multi-task learning loss. For multi-task learning, two separate projection heads for both $\mathcal{L}_{c}$ and cross entropy loss $\mathcal{L}_{ce}$ are simultaneously trained. $\mathcal{L}_{ce}$ is calculated using the output predictions of a classifier module $C_\eta : {\mathcal{H}} \to \mathcal{\hat{Y}}$. As labels are a requirement for calculating $\mathcal{L}_{ce}$, $D_{\text{ANNO}}$ is employed. For computing the loss for multi-task contrastive learning, we combine the loss from two separate projections:

\begin{equation}
    \mathcal{L}_{multi} = \alpha_{c}\mathcal{L}_{c} + \alpha_{ce}\mathcal{L}_{ce}
\label{eq:multi}
\end{equation}

where $\alpha_{c}$ and $\alpha_{ce}$ are the scaling factors for the unsupervised contrastive and supervised cross entropy losses, respectively.

\subsection*{Prognostic Outcome Classification via Multiple Instance Learning} \label{sec:mil}
Multiple instance learning (MIL) is a suitable approach for prognostic classification due to its ability to handle inherent uncertainties, diversity and intricacies present in medical data \cite{dietterich1997solving}. A dataset $\mathcal{H}, \mathcal{Y} = \left \{ (\mathbf{H}^i,y^i), \forall i=1,...,N \right \}$ is formed of pairs of bag instances $\mathbf{H}$ and their corresponding labels $y$, where $i$ denotes the current sample for a total of $N$ samples. A bag $\mathbf{H}$ consists of instances $\mathbf{h}_l$:

\begin{equation}
    \mathbf{H}= \left \{ \mathbf{h}_l, \forall l=1,...,L \right \}
    \label{eq:boi}
\end{equation}

where $L$ is the number of instances in the bag.  Among MIL model architecture variants, we adopted attention-based multiple instance learning (AbMIL). Given a label for a patient, a model should infer which ROIs visual features lead to predicting the patient’s prognostic outcome. An attention score $a_i$ for a feature embedding $\mathbf{h}_i$ can be calculated as:

\begin{equation}
    a_{i} =  \frac{\mathrm{exp}\{ \mathbf{w}^{\top} (\mathrm{tanh}(\mathbf{Vh_{i}^{\top}}) \odot \mathrm{sigm}(\mathbf{Uh_{i}^{\top})})\}} {\sum_{l=1}^{L}\mathrm{exp}\{ \mathbf{w}^{\top} (\mathrm{tanh}(\mathbf{Vh_{l}^{\top}}) \odot \mathrm{sigm}(\mathbf{Uh_{l}^{\top})}) \}}
\label{eq:attention}   
\end{equation}

where $\mathbf{w}\in\mathbb{R}^{L\times 1}$, $\mathbf{V}\in\mathbb{R}^{L\times M}$ and $\mathbf{U}\in\mathbb{R}^{L\times M}$ are trainable parameters and $\odot$ is an element-wise multiplication. Furthermore, the hyperbolic tangent $\mathrm{tanh}(\cdot)$ and sigmoid $\mathrm{sigm}(\cdot)$ are included to introduce non-linearity for learning complex applications. The strength of the attention modules is not only in terms of interpretability, but also in predictive power, as attention scores are directly influencing the forward propagation of the model. Once the attention scores are obtained, we obtain the patient prediction $\hat{y}$ as:

\begin{equation}
    \hat{y} = \Psi_\rho(\mathbf{A} \cdot \mathbf{H})
\end{equation}

where $\Psi_\rho$ is a MLP acting as a patient classifier. Additionally, we propose using NMIA \cite{fuster2022nested}. NMIA defines a bag can consisting of multiple sub-bags, which contain the instances themselves. This serves to further stratify into clusters or regions, and accurately represent the arrangement of the scattered data, where tiles belong to particular tissue areas and these themselves to the WSI. A bag-of-bags for a WSI $\mathbf{H}_{\text{WSI}}$ contains a set of inner-bags, or regions, $\mathbf{H}_{\text{REG},k}$:
\begin{equation}
    \mathbf{H}_{\text{WSI}} = \left \{ \mathbf{H}_{\text{REG},k}, \forall k=1,...,K \right \}
    \label{eq:bob}
\end{equation}

where the number of inner-bags $K$ varies between different  WSI. Ultimately, $\mathbf{H}_{\text{REG},k}$ contain instance-level representations $\mathbf{h}_{\text{TILE},l}$ of tiles located within the physical region.

For this project, we explored three configurations: using image data, using clinicopathological data and fusing both. For image data, we applied a weakly supervised architecture. For clinical data, we sorted it into a 1-dimensional vector $\mathbf{h}_{var}$ and fed it directly to the patient classifier $\Psi_\rho$. As for the combination of both, we used  a weakly supervised architecture for generating the patient embedding representation and concatenated the clinical features to said embedding as $\mathbf{h}_{cli} = [\mathbf{A} \cdot \mathbf{H}, \mathbf{h}_{var}]$.

\section*{Experimental Setup}

In this section, we present the carefully designed experiments. 
First, to evaluate the proposed pipeline structure, we present experiments on artificial data as well as an application with region based labels. 
From there on, we consider two prognostic applications: BCG response prediction and recurrence prediction.  Hyperparameters, like the choice of optimizer, loss, and others are identical in all experiments. The code is available at \href{https://github.com/Biomedical-Data-Analysis-Laboratory/HistoPrognostics}{\textcolor{magenta}{https://github.com/Biomedical-Data-Analysis-Laboratory/HistoPrognostics}}. 

\subsubsection*{ROI Extraction} The fully automatic tissue segmentation was performed according to \cite{wetteland2020multiscale} for recurrence $S_{\text{SUH}}$ and \cite{fuster2023active} for BCG treatment $S_{\text{EMC}}$. Different regions were extracted as explained in Section \nameref{sec:segmentation}.

\subsubsection*{Contrastive Feature Representations}
We define a temperature parameter to 0.07 for the calculation of the loss for contrastive learning, as explained in Section \nameref{sec:feature}. We experiment with different CNN backbones;  VGG16, DenseNet121 and ResNet18, with initial  weights $\theta_{I}$, from pretraining on ImageNet \cite{krizhevsky2012imagenet}. For the supervised approaches, labels of grading and presence of TILs were used as diagnostic factors. In order to weigh the impact of the training size, we also run experiments of the unsupervised variant with larger, but limited, training samples. With respect to multi-task learning of unsupervised contrastive and cross entropy classification, we set the parameters $\alpha_c$ and $\alpha_{ce}$ to 1.0 and 0.5, respectively. Adam was set as optimizer with a learning rate of 1e-4, a batch size of 128, and a total of fixed 10 epochs. The augmentations applied consisted of flip, flop, rotation, affine transformations, and color jittering.

\subsubsection*{Prognosis Classification} 
Bladder cancer recurrence is indeed regarded as a manifestation of treatment failure, as it indicates that the initial treatment did not effectively eradicate all cancer cells. Building upon this rationale, we will employ the data set $S_{\text{EMC}}$, see Table \ref{tab:patients}, in our decision-making process regarding the selection of feature extractors, contrastive loss functions, ROI selection, and magnification levels for both prognostic applications. This choice derives from the larger number of patients in the dataset, being a more representative sample of the population under study. Focal Tversky loss (FTL) is employed \cite{abraham2019novel}. FTL employs two parameters, denoted as $\alpha_{l}$ and $\gamma_{l}$, which allow for adjusting the focus on different classes and handling the difficulty of training examples, respectfully. We also set an early stopping criteria of 30 epochs based on the AUC score on the validation set. We also implement a 5-runs Montecarlo with a 5\% dropout for sampling purposes. A grid hyperparameter search is done to find the optimal values for the given task. The search includes bag sampling $n_{b}$, learning rate $lr$, optimizer $opt$, dropout rate $d_r$, number of neurons in the classifier $n_{\Theta_\rho}$ and attention mechanism $n_{att}$, loss functions parameters $\alpha_{l}$ and $\gamma_{l}$. The list of hyperparameters with their corresponding possible values and the resulting choice can be seen in Table \ref{tab:hyper}:

\begin{table}[h]
\caption{\label{tab:hyper} \textbf{Results of hyperparameter search for optimizing predictive performance.}}
\centering
\begin{tabular}{|c|c|}
\hline
\textbf{Hyperparameter}                                              & \multicolumn{1}{c|}{\textbf{List of values}} \\ \hline \hline
$lr$          &                 $10^{-1}, 10^{-2}, 10^{-3}, 10^{-4}$                             \\ \hline
$opt$               &                   SGD, Adam                           \\ \hline
$n_{b}$     &                     4, 16, 64, 256, $L$                         \\ \hline
$d_r$           &                         0.1, 0.2, 0.3, 0.4, 0.5                     \\ \hline
$\alpha_{l}$         &                         0.0, 0.3, 0.6, 0.9                     \\ \hline
$\gamma_{l}$           &                            0.5, 1, 2                  \\ \hline
$n_{\Theta_\rho}$ &                         128, 512, 1024, 4096                      \\ \hline
$n_{att}$   &                         128, 512, 1024, 4096                     \\ \hline \hline
\makecell{\textbf{Hyperparameter} \\ \textbf{choice}}   &   \makecell{$lr=10^{-2}, opt=$SGD$, n_b=64$ \\ $d_r=0.2, \alpha_l=0.9, \gamma_l=2.0,$ \\ $n_{\Theta_\rho}= \left\{4096,2048 \right\}, n_{att}= 4096$}                        \\ \hline
\end{tabular}
\end{table}

\section*{Preliminary Experimentation}
The three-step pipeline is convoluted, and prognostic labels are weak, typically involving one label per patient. To evaluate the pipeline, we conduct experiments in more controlled settings. Firstly, we aim to assess the pipeline using synthetic data generated from distributions with varying degrees of overlap. Thereafter, we evaluate the pipeline on actual WSI data, focusing on a diagnostic task with strong region-based labels, specifically the detection of regions containing lymphocytes.

\subsection*{Effects of Different Data Distributions}
We define three different matching distributions pairs, which can be described as distributions with minor, partial overlap, or significant overlap. The degree of overlap is tuned using the mean $\mu$ and standard deviation $\sigma$ parameters. For all experiments, we define $\mathcal{P}_{3,2.5}$ as our class 0. For the matching distribution of class 1, we use $\mathcal{P}_{-3,1}$, $\mathcal{P}_{0,2}$ and $\mathcal{P}_{2,1.5}$ for minor, partial and significant overlap, respectively. A total of 150 bags are created, balanced among two classes, with a 90, 30, 30 split for train, validation and test, respectively. The number of instances per bag is variable $N \in [3000, 7000]$, without replacement. The number of positive instances is limited to be less than half.


Results indicate that it is uncomplicated to discern between $\mathcal{P}_{-3,1}$ and $\mathcal{P}_{0,2}$ distributions, as shown in Table \ref{tab:prob}. However, we observe that for an overlapping distribution $\mathcal{P}_{2,1.5}$ the learning breaks down. We wanted to go further into finding the breaking point for the partial overlap distribution $\mathcal{P}_{0,2}$, and as such, we tried different parameters regarding the bag label balance in the train set and the representation of positive samples in bags. We look into a percentage of positive bags of 25 and 40\%, as well as the number of positive samples in the range of 0 to 50\%, and 0 to 75\%. Results reveal the significance of class imbalance in bag classification, and the representation of positive and negative samples within the bags. The classification of highly imbalanced bag datasets can be challenging. While augmenting the number of positive instances can enhance the instance classification performance, it is insufficient to achieve satisfactory bag classification results. The performance naturally improves as the bag distribution becomes more balanced, but it is not enough unless the positive class is over-represented.

\begin{table}[h]
\caption{\textbf{Prediction probabilities for discerning $\mathcal{P}_{3,2.5}$.}}
\centering
\begin{tabular}{|c|c|}
\hline
    \textbf{Class 1} & \textbf{AUC} \\ \hline \hline
    $\mathcal{P}_{-3,1}$ & \textbf{1.000} \\ \cline{1-1}
    $\mathcal{P}_{0,2}$ & \textbf{1.000} \\ \cline{1-1}
    $\mathcal{P}_{2,1.5}$ & 0.500 \\ \hline
\end{tabular}
\label{tab:prob}
\end{table}

\subsection*{Detection of Lymphocytes}
We evaluated the performance of our proposed solution in a diagnostic task using the same WSI as those used in $S_{\text{EMC}}$. Specifically, we defined a problem of detecting the presence of tumor-infiltrating lymphocytes (TILs), immune cells which have been associated with improved patient outcomes \cite{kamat2016definitions}. Tiles of size 256 × 256 were extracted from annotated lamina propria areas, with and without TILs, at 40x magnification. As a pre-processing step, tiles were normalized and resized to 224 × 224. Augmentations performed consisted of rotation and jittering operations.
We define two experiments for different percentages of tiles with stromal TILs. The first one, $MIL_{1+}$, follows the standard MIL convention of detecting at least one tile with TILs. As for the second one, $MIL_{t+}$, we define a threshold $t$ for which more than 50\% of the tiles must contain TILs.

We employed $MIL_{1+}$ to determine the optimal configuration for feature extraction. The results are presented in Table \ref{tab:til_feature}. Pre-trained features from natural images indeed capture relevant features for histopathological classification. However, the unsupervised approach $\theta_{\text{C}}$ exhibits superior performance in bag-level classification.  $MIL_{t+}$ is therefore only tested with $G_{\theta_{C}}$, and this provides the best result with an AUC of 1.0. 
This might be due to the fact that labels are noisy, as many of the tiles annotated TIL-free actually contain some, even if the count is low. Using $MIL_{t+}$ makes it easier for the model to find a threshold $t$ than an absolute presence or absence given the noisy weak labels. Consistent with the observations on synthetic data, bags containing a substantial number of positive instances exhibit superior performance. This implies that accurate predictions rely on the existence of several positive instances within a bag, with less emphasis on the presence of positive instance outliers.

\begin{table}[h]
\caption{\textbf{TIL detection comparison for different feature extractors.}}
\centering
\begin{tabular}{|c|c|c|}
\hline
    \textbf{MIL} & \textbf{Weights $\theta$}     & \textbf{AUC} \\ \hline \hline
    \multirow{5}{*}{$MIL_{1+}$} & $\theta_{I}$ & 0.833 \\ \cline{2-2}
     & $\theta_{\text{CE}}$ & 0.938 \\ \cline{2-2}
     & $\theta_{\text{SC}}$ & 0.909 \\ \cline{2-2}
     & $\theta_{\text{C}}$ & 0.940 \\ \cline{2-2}
     & $\theta_{\text{MULTI}}$ & 0.938 \\ \hline \hline
    $MIL_{t+}$ & $\theta_{\text{C}}$ & \textbf{1.000} \\ \hline
\end{tabular}
\label{tab:til_feature}
\end{table}

\section*{Prognostic Experiments}
In this section, we utilize the proposed three-step pipeline for two prognostic applications: BCG response prediction on the $S_{\text{EMC}}$ dataset, and recurrence prediction on the $S_{\text{SUH}}$ dataset. Given the numerous choices for contrastive learning loss, backbone, ROI, and magnification levels, coupled with the computational intensity of learning and inference on gigapixel images, we adopt a systematic experiment approach. Therefore, we focus on testing one factor at a time and restrict further testing to the most promising results. In order to identify the optimal choices, we couple an AbMIL classification module to assess the resulting classification performance, using the validation subset. In subsections \nameref{chap:fe} to \nameref{chap:comp}, we use BCG response prediction with $S_{\text{EMC}}$ as the reference task, while recurrence prediction with $S_{\text{SUH}}$ is discussed in subsection \nameref{chap:comp}. Finally, subsection \nameref{chap:att} discusses the interpretability of trained models.

\subsection*{Feature Extraction and Contrastive Learning}
\label{chap:fe}
We aim to select a CNN backbone of preference for the feature extractor $G_\theta$, referring to the second step in the pipeline. The most commonly used backbones in CPATH literature are DenseNet, ResNet and VGG \cite{cui2021artificial}. To ensure a fair performance comparison in a contrastive learning approach, we will incorporate the use of different labels, which will either be unsupervised, supervised contrastive or multi-task learning. 
In this experiment, we constrict the ROI to be $D_{\text{URO}}^{20x}$, with patches extracted at 20x. The urothelium tissue type is widely recognized as highly informative, and the choice of 20x magnification strikes a balance between capturing morphological structure context and preserving cellular details. 

Results for classification can be found in Table \ref{tab:contrastive_fe}. 
DenseNet121 offers promising results in terms of AUC performance for classification, without compromising computational efficiency. Therefore, we will use DenseNet121 as the backbone for the remaining experiments. The best performing contrastive learning strategy is unsupervised $\mathcal{L}_{c}$. Hence, we will proceed to use the frozen weights from the unsupervised method $\theta_{\text{C}}$.

\begin{table}[h]
\caption{\label{tab:contrastive_fe} \textbf{Validation AUC scores for $D^{20x}_{\text{URO}}$ BCG.}}
\centering
\begin{tabular}{|c|c|c|c|}
\hline
\textbf{Weights $\theta$}                                  & \textbf{DenseNet121} & \textbf{ResNet18} & \textbf{VGG16} \\ \hline \hline
$\theta_{I}$   &     0.576(0.029)              &      0.474(0.009)           &      0.549(0.005)        \\ \cline{1-1}
$\theta_{\text{C}}$ &      \textbf{0.672(0.032)}             &         0.628(0.017)         &      0.506(0.022)         \\ \cline{1-1}
$\theta_{\text{SC}}$   &       0.521(0.054)            &        0.568(0.009)         &       0.480(0.036)       \\ \cline{1-1}
$\theta_{\text{MULTI}}$   &        0.515(0.031)           &     0.506(0.042)             &       0.434(0.048)        \\    \hline
\end{tabular}
\caption*{\textit{The results show the mean and standard deviation over 5 runs.}}
\end{table}

\subsection*{Region of Interest Selection}
After determining the CNN backbone and contrastive learning strategy, the next step involves identifying the ROI with the best discriminative capability for the prognostic task. As highlighted in Fig. \ref{fig:datasets}, the ROIs are found from either manual annotations or from the output of the automated tissue segmentation model. From the automatically segmented regions, we obtain $D_{\text{URO}}^{20x}$, $D_{\text{LP}}^{20x}$, $D_{\text{UROLP}}^{20x}$, $D_{\text{BORDER}}^{20x}$, $D_{\text{FRONT}}^{20x}$, while $D_{\text{ANNO}}^{20x}$ is generated from the annotated set. 

\begin{figure}[h]
\centerline{\includegraphics[height=0.5\columnwidth, trim={0.75cm 0cm 1cm 1.7cm},clip]{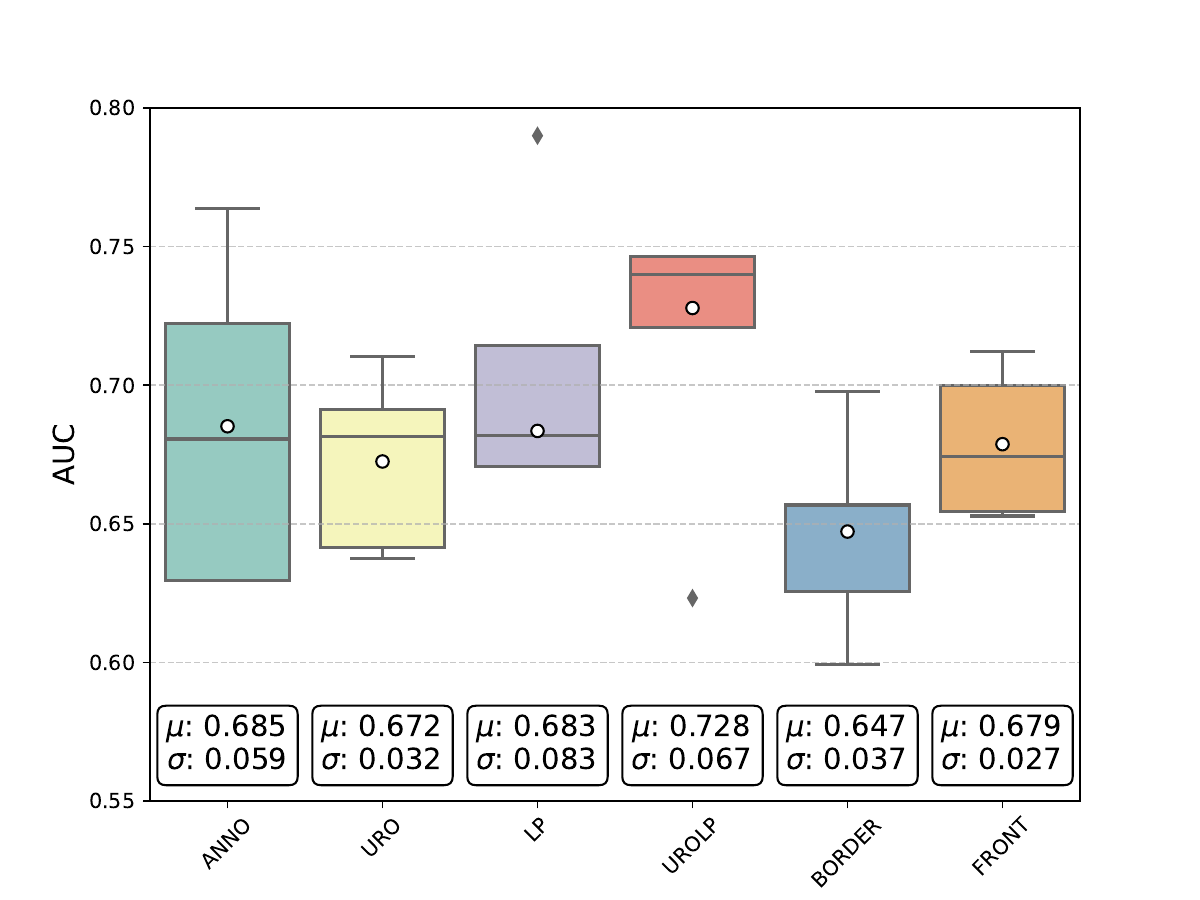}}
\caption{\textbf{Box plot illustrating AUC performance variation across validation sets with different ROIs for 20x magnification.} Notably, $D^{20x}_{\text{UROLP}}$ emerges as the top performer amongst the ROIs. White dots represent the average value $\mu$, and black diamonds represent outliers. The results show the mean $\mu$ and standard deviation $\sigma$ over 5 runs.}
\label{fig:box}
\end{figure}

A comparison is shown in Fig \ref{fig:box}. The comparison underscores $D^{20x}_{\text{UROLP}}$'s advantage, showcasing its consistent and balanced classification performance with an average AUC score of 0.728. That said, $D^{20x}_{\text{FRONT}}$ has the lowest variance across runs, reaching the most consistent results. $D^{20x}_{\text{ANNO}}$ demonstrates the second-highest average AUC score, but the highest variance. This implies a promising characteristic for future development of diagnostic models. Emulating a pathologist sense of expertise, a model could automatically identify diagnostically relevant areas and generate annotations. Then, these generated annotations could be used for further enhancing feature discrimination for subsequent prognostic models. That said, in our investigation, only a limited set of tiles is annotated. This results in a mismatch in training size between datasets automatically generated and annotations. Moreover, an independent system cannot rely on expert input during the inference stage. Moving forward, we take all three aforementioned ROIs for determining the optimal magnification input choice.

\subsection*{Magnification Level Choice}
We aim to assess the influence of varying magnification levels on the performance, testing mono-scale magnification levels of 10x, 20x and multi-scale TRI (2.5x, 10x, 40x).
The results are shown in Table \ref{tab:mag}, and allows us to assess the model's accuracy in capturing both broader tissue context and finer cellular details, as well as contextual neighbouring regions as per the multi-scale input.

\begin{table}[h]
\caption{\label{tab:mag} \textbf{Validation AUC scores for BCG at different ROIs and magnification levels.}}
\centering
\begin{tabular}{|c|c|c||c|}
\hline 
\textbf{Magnification}                                   & \textbf{UROLP}  &  \textbf{FRONT}  &  \textbf{ANNO}* \\ \hline \hline
\textbf{10x}   &        0.621(0.021)  &        0.647(0.033)         &       \textbf{0.790(0.113)}    \\ \cline{1-1}
\textbf{20x}   &       \textbf{0.728(0.067)}   &        0.679(0.027)         &       0.685(0.059)     \\ \cline{1-1}
\textbf{TRI}   &         0.649(0.014)     &         0.621(0.038)         &       0.615(0.029)   \\ \cline{1-1}  \hline
\end{tabular}
\caption*{\textit{The results show the mean and standard deviation over 5 runs. \\ * Not all train and validation WSI were annotated.}}
\end{table}

Utilizing a fixed magnification level throughout the image consistently yields AUC levels that hover around 0.7 at 20x magnification, the highest for the automatic ROI $D^{20x}_{\text{UROLP}}$. However, it's worth noting that the predictive performance for 10x is worse. For comparison, $D^{10x}_{\text{ANNO}}$ obtained on average higher AUC values, yet, constrained to expert input and lack of a complete set of annotations. 
Despite of its capacity to capture structural intricacies, the multi-scale model TRI stands with lower performance. TRI performs worse possibly because the larger input requires more training data for generalization.

\subsection*{Weakly Supervised Aggregation for Treatment Outcome Prediction}
We conduct a comparison between five aggregation techniques in weakly supervised learning. These include majority voting, max, mean, AbMIL and NMIA. These techniques are evaluated to determine their effectiveness in combining and summarizing information from multiple instances. 

In Table \ref{tab:agg}, we present the obtained results. Among these techniques, those involving majority voting, mean and max aggregation, which do not include in-built attention mechanisms, demonstrate less promising outcomes. In contrast, the utilization of AbMIL leads to a notable performance improvement. Moreover, when examining the scattered tissue regions across the WSI using the nested multiple instance method, NMIA, we observe the most significant performance enhancement across all techniques with an AUC of 0.678.

\begin{table*}[h]
\begin{adjustwidth}{-2.25in}{0in}
\caption{\label{tab:agg} \textbf{Test AUC performance for $D^{20x}_{\text{UROLP}}$ for BCG and recurrence.}}
\centering
\begin{tabular}{|c|c|c|c||c|c|c|}
\hline
\multirow{2}{*}{\textbf{Method}} & \multicolumn{3}{c||}{\textbf{BCG}} & \multicolumn{3}{c|}{\textbf{Recurrence}}\\
\cline{2-7}
                                   & \textbf{Best Run}  &       \textbf{$\mu(\sigma)$}    &       \textbf{Montecarlo} & \textbf{Best Run}  &       \textbf{$\mu(\sigma)$}    &       \textbf{Montecarlo}\\ \cline{1-1}
\hline \hline
\textbf{Maj. Voting}   &       0.500        &       0.497(0.003)     &       0.440(0.075) &       0.421        &       0.420(0.001)     &       0.420(0.001)\\ \cline{1-1}
\textbf{Mean}   &       0.427        &       0.417(0.007)    &       0.446(0.033) &       0.615        &       \textbf{0.583(0.023)}    &       0.613(0.015)\\ \cline{1-1}
\textbf{Max}   &      0.520         &       \textbf{0.499(0.011)}    &       0.487(0.007) &      0.555         &       0.517(0.047)    &       0.544(0.008)\\ \cline{1-1}
\textbf{AbMIL}   &        0.549       &       0.434(0.077)     &       0.548(0.007) &        0.592       &       0.500(0.064)     &       0.593(0.002)\\ \cline{1-1}
\textbf{NMIA}   &      \textbf{0.678}         &       0.486(0.111)     &      \textbf{0.612(0.009)} &      \textbf{0.721}         &       0.580(0.079)     &      \textbf{0.722(0.003)}  \\ \cline{1-1} \hline \hline
\textbf{Clinical}   &      0.501        &       0.471(0.024)     &       0.498(0.017) &     -        &       -     &       -\\ \cline{1-1}
\textbf{Clinical + AbMIL}   &    0.502        &       0.456(0.037)     &       0.491(0.006) &    -        &       -     &       -\\ \cline{1-1}
\textbf{Clinical + NMIA}   &    0.475        &       0.450(0.036)     &       0.454(0.008) &    -        &       -     &      -\\ \hline

\end{tabular}
\caption*{\textit{The results show the mean and standard deviation over 5 runs.}}
\end{adjustwidth}
\end{table*}

\subsection*{Fusing Image and Clinicopathological Data}
By fusing clinicopathological and image data with the aim of complementing each other, as explained in subsection \nameref{sec:mil}, deep learning models should gain a comprehensive understanding of patient conditions. However, it is imperative to acknowledge the potential limitations and challenges inherent in each form of data. While images offer visual cues that might be ambiguous without clinical context, reports may lack the granularity and specificity present in visual data. Clinicopathological data is often compiled from various sources, including physician notes, laboratory test results, and imaging, potentially introducing noise and variability. Conversely, WSI may also contain inherent variability due to factors such as staining variations or tissue preparation techniques.

In our experiments, unexpectedly, the findings displayed in Table \ref{tab:agg} indicate that histological features may offer more pertinent information for predicting patient outcomes compared to either clinical data alone or their fusion. This finding raises questions about the traditional reliance on clinical data and highlights the potential of histopathological images as a standalone predictor for improved prognostic accuracy. Moving forward, exploring different fusion methods for integrating clinicopathological and image data could yield insights into the optimal approach for maximizing predictive performance. Moreover, incorporating additional clinical parameters may provide a more comprehensive understanding of the status of the disease.

\subsection*{On the Importance of Manual Annotations}
\label{chap:comp}
At the inference stage, a fully-independent system cannot be conditioned by manually annotated regions. Even at the training stage, we have observed the challenges in obtaining annotations for all WSI due to the labor-intensive nature of the process. For a prognostic task, the relevance of manual annotations remains uncertain. To explore this aspect, we conducted an experiment where the model was trained on annotated regions ANNO, but tested on automatically segmented regions AUTO, and vice versa. This was compared to a fully automated system for both training and inference.
Table \ref{tab:ablation} shows the results, where the ROI-column indicates which ROI definition was used for the AUTO set. For comparison, using ANNO for both train and test gives a performance of 0.441(0.031). With one exception, the models perform better after using AUTO-generated ROIs in the training and ANNO for testing, which we interpret as the models benefiting from the larger dataset that is available when we can include non-annotated data.

\begin{table*}[h]
\begin{adjustwidth}{-2.25in}{0in}
\caption{\label{tab:ablation} \textbf{Comparing AUC performance for distinct training and test datasets, for $S_{\text{EMC}}$ 20x magnification.}}
\centering
\begin{tabular}{cc||c|c|c|c|c|}
\hline
\multicolumn{1}{|c|}{\textbf{ROI (Train/Test)}}                      &   \textbf{Architecture}  & \textbf{URO} & \textbf{LP} & \textbf{UROLP} & \textbf{BORDER} & \textbf{FRONT} \\ \hline \hline
\multicolumn{1}{|c|}{\multirow{2}{*}{\textbf{ANNO / AUTO}}} & \multicolumn{1}{|c||}{\textbf{AbMIL}} &       0.489(0.032)          &     0.446(0.055)         &         \textbf{0.528(0.026)}           &     0.463(0.054)            &      0.408(0.025)         \\ 
\multicolumn{1}{|c|}{}  &  \multicolumn{1}{|c||}{\textbf{NMIA}}  &       0.543(0.022)         &     0.476(0.054)         &        0.475(0.032)           &     0.468(0.054)            &      0.503(0.027)         \\ \cline{1-1} \hline \hline
\multicolumn{1}{|c|}{\multirow{2}{*}{\textbf{AUTO / ANNO}}} &  \multicolumn{1}{|c||}{\textbf{AbMIL}}  &      \textbf{0.611(0.067)}             &       0.468(0.144)          &      0.508(0.090)         &           \textbf{0.600(0.094)}         &      \textbf{0.663(0.098)}        \\ 
\multicolumn{1}{|c|}{} &  \multicolumn{1}{|c||}{\textbf{NMIA}}  &      0.486(0.119)             &        \textbf{0.601(0.082)}          &      0.519(0.054)         &          0.525(0.054)         &      0.531(0.073)        \\ \cline{1-1} \hline \hline
\multicolumn{1}{|c|}{\multirow{2}{*}{\textbf{AUTO / AUTO}}} &  \multicolumn{1}{|c||}{\textbf{AbMIL}}  &      0.492(0.065)             &       0.452(0.060)          &      0.434(0.077)         &          0.534(0.036)         &      0.484(0.066)        \\ 
\multicolumn{1}{|c|}{} &  \multicolumn{1}{|c||}{\textbf{NMIA}}  &      0.535(0.046)             &       0.513(0.043)          &      0.486(0.111)         &       0.512(0.053)      &   0.428(0.028)     \\ \cline{1-1} \hline
\end{tabular}
\caption*{\textit{The results show the mean and standard deviation over 5 runs.}}
\end{adjustwidth}
\end{table*}

\subsection*{Recurrence Prediction}
\label{chap:rec}
Following the predefined settings for ROI, magnification, and feature extraction used in the BCG application, we apply the same configuration for recurrence prediction. 
The results are presented in Table \ref{tab:agg}.
We observe that the utilization of mean aggregation demonstrated the highest average predictive performance. Nevertheless, NMIA exhibited the most substantial performance enhancement among all the techniques studied, with an AUC of 0.721. This underscores the effectiveness of incorporating nested attention mechanisms for the accurate identification of crucial patterns within tissue regions. 

\subsection*{Attention-guided Interpretability}
\label{chap:att}
The attention scores obtained at the inference stage can be useful for interpretability reasons and can give knowledge on what the model considers relevant for a given prediction. Instances with higher attention scores often correspond to pivotal regions pertinent to the predicted label, aiding in both model validation and reasoning. 
Attention scores aid pathologists in understanding model's rationale, highlighting areas of interest. This interpretability is valuable for either positive or negative predictions, facilitating validation and refinement of the model's decisions. 

\begin{figure*}[h]
\centerline{\includegraphics[width=\linewidth]{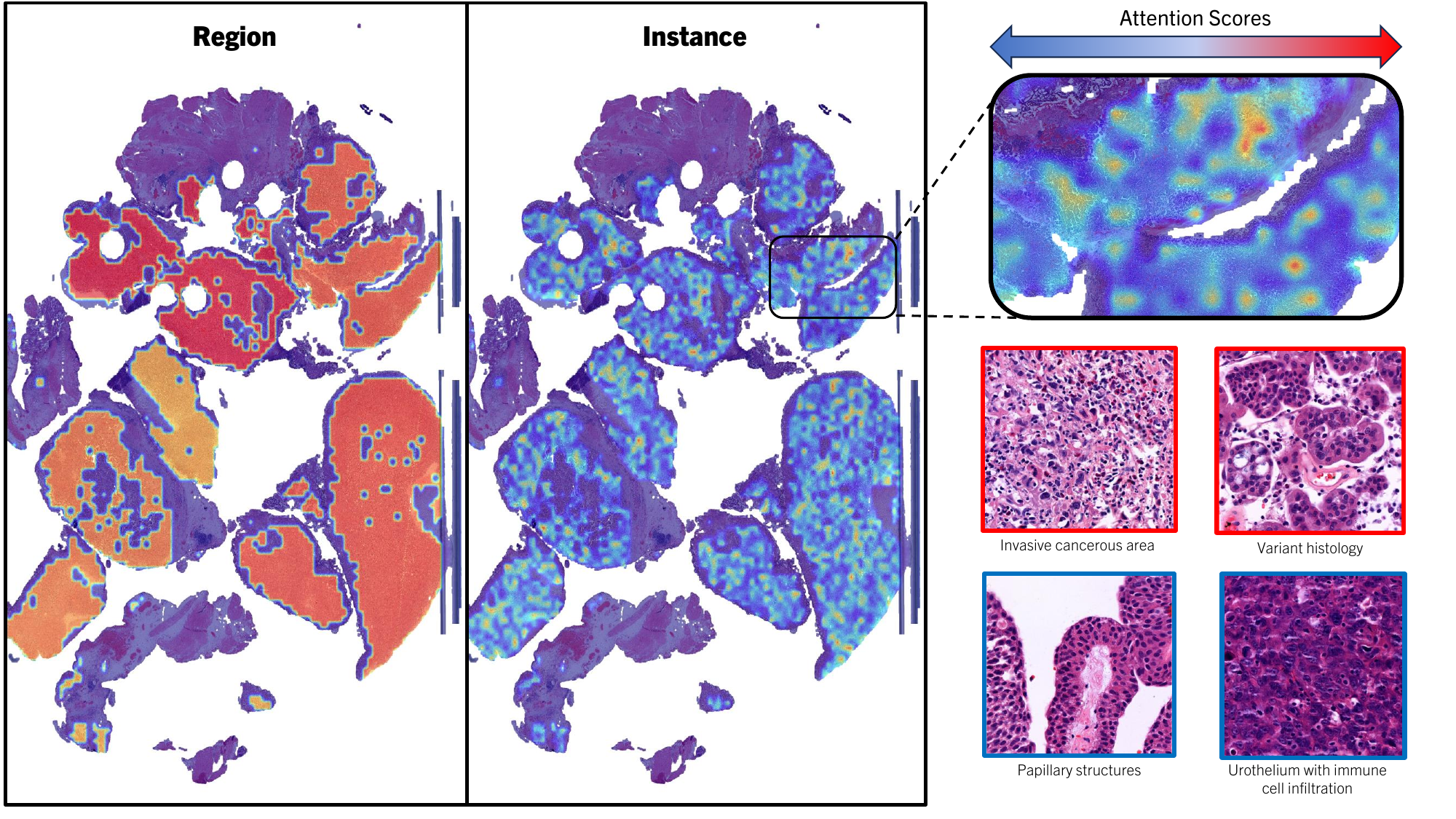}}
\caption{\textbf{Heatmap illustrating attention scores over a BCG-NR WSI from $S_{\text{EMC}}$.} The heatmap provides insights into the ROIs where the attention is concentrated within the WSI, facilitating a better understanding of prediction dynamics and highlighting areas of significance for clinical interpretation.}
\label{fig:attentionmap}
\end{figure*}

An example of attention score heatmap is visualized in Fig. \ref{fig:attentionmap}. 
Pathologists often use tissue punching to select clinically relevant regions for subsequent analysis in tissue microarrays. While the model consistently allocates its highest attention to punched areas, it is crucial to acknowledge that diagnostically relevant information may extend beyond the punched areas.  Nevertheless, the alignment between the model's attention focus and the clinical practice of region selection through punching is an encouraging and promising finding.

\section*{Conclusion}
We propose a three-step automated pipeline for the challenging task of prognostic prediction, eliminating the need for manual annotations. This trend is expected to persist in the future of CPATH, given the numerous tasks and limited annotation resources available.

The process begins with automated ROI segmentation, where task-specific knowledge can be combined with an automatic tissue classifier to extract relevant ROI. Since prognostic labels are weak in nature, the second step employs contrastive learning to train the feature extractor. Finally, an attention-based nested multiple instance learning classifier, providing predictions and insights into the crucial regions of the image.
The pipeline demonstrates exceptional performance on a simpler diagnostic task, achieving a perfect score with an AUC of 1.0 for detecting areas with tumor-infiltrating lymphocytes. 
Additionally, experiments on synthetic data confirm that the pipeline works perfectly when input distributions are distinct. However, performance drops when they become highly overlapping. 
In our study, we conducted a thorough pioneering investigation into the use of deep learning and histopathological images for prognostic prediction. We employ response to BCG treatment in patients with HR-NMIBC and recurrence in NMIBC patients as uses cases. The utilization of deep learning for BCG response prediction using WSI is a first.
The most promising outcomes reveal AUC values of 0.678 for BCG outcome prediction and 0.721 for recurrence prediction. While there is room for improvement, achieving fully automated prognostics based solely on WSI remains a challenging task.

\section*{Acknowledgments}
This research has received funding from the European Union's Horizon 2020 research and innovation program under grant agreements 860627 (CLARIFY).

\nolinenumbers

%
%
%


\begin{thebibliography}{10}

\bibitem{van2021deep}
Van~der Laak J, Litjens G, Ciompi F.
\newblock Deep learning in histopathology: the path to the clinic.
\newblock Nature medicine. 2021;27(5):775--784.

\bibitem{madabhushi2016image}
Madabhushi A, Lee G.
\newblock Image analysis and machine learning in digital pathology: Challenges and opportunities.
\newblock Medical image analysis. 2016;33:170--175.

\bibitem{babjuk2022european}
Babjuk M, Burger M, Capoun O, Cohen D, Comp{\'e}rat EM, Escrig JLD, et~al.
\newblock European Association of Urology guidelines on non--muscle-invasive bladder cancer (Ta, T1, and carcinoma in situ).
\newblock European urology. 2022;81(1):75--94.

\bibitem{fuster2022nested}
Fuster S, Eftest{\o}l T, Engan K.
\newblock Nested multiple instance learning with attention mechanisms.
\newblock In: 2022 21st IEEE International Conference on Machine Learning and Applications (ICMLA). IEEE; 2022. p. 220--225.

\bibitem{burger2013epidemiology}
Burger M, Catto JW, Dalbagni G, Grossman HB, Herr H, Karakiewicz P, et~al.
\newblock Epidemiology and risk factors of urothelial bladder cancer.
\newblock European urology. 2013;63(2):234--241.

\bibitem{kamat2016definitions}
Kamat AM, Sylvester RJ, B{\"o}hle A, Palou J, Lamm DL, Brausi M, et~al.
\newblock Definitions, end points, and clinical trial designs for non--muscle-invasive bladder cancer: recommendations from the International Bladder Cancer Group.
\newblock Journal of Clinical Oncology. 2016;34(16):1935.

\bibitem{cui2021artificial}
Cui M, Zhang DY.
\newblock Artificial intelligence and computational pathology.
\newblock Laboratory Investigation. 2021;101(4):412--422.

\bibitem{kourou2015machine}
Kourou K, Exarchos TP, Exarchos KP, Karamouzis MV, Fotiadis DI.
\newblock Machine learning applications in cancer prognosis and prediction.
\newblock Computational and structural biotechnology journal. 2015;13:8--17.

\bibitem{zheng2022accurate}
Zheng Q, Yang R, Ni X, Yang S, Xiong L, Yan D, et~al.
\newblock Accurate Diagnosis and Survival Prediction of Bladder Cancer Using Deep Learning on Histological Slides.
\newblock Cancers. 2022;14(23):5807.

\bibitem{tokuyama2022prediction}
Tokuyama N, Saito A, Muraoka R, Matsubara S, Hashimoto T, Satake N, et~al.
\newblock Prediction of non-muscle invasive bladder cancer recurrence using machine learning of quantitative nuclear features.
\newblock Modern Pathology. 2022;35(4):533--538.

\bibitem{de2023non}
de~Jong FC, Laajala TD, Hoedemaeker RF, Jordan KR, van~der Made AC, Boev{\'e} ER, et~al.
\newblock Non--muscle-invasive bladder cancer molecular subtypes predict differential response to intravesical Bacillus Calmette-Gu{\'e}rin.
\newblock Science Translational Medicine. 2023;15(697):eabn4118.

\bibitem{wang2019weakly}
Wang X, Chen H, Gan C, Lin H, Dou Q, Tsougenis E, et~al.
\newblock Weakly supervised deep learning for whole slide lung cancer image analysis.
\newblock IEEE transactions on cybernetics. 2019;50(9):3950--3962.

\bibitem{zhou2021histopathology}
Zhou C, Jin Y, Chen Y, Huang S, Huang R, Wang Y, et~al.
\newblock Histopathology classification and localization of colorectal cancer using global labels by weakly supervised deep learning.
\newblock Computerized Medical Imaging and Graphics. 2021;88:101861.

\bibitem{wang2022weakly}
Wang CW, Chang CC, Lee YC, Lin YJ, Lo SC, Hsu PC, et~al.
\newblock Weakly supervised deep learning for prediction of treatment effectiveness on ovarian cancer from histopathology images.
\newblock Computerized Medical Imaging and Graphics. 2022;99:102093.

\bibitem{andreassen2023deep}
Andreassen C, Fuster S, Hardardottir H, Janssen EAM, Engan K.
\newblock Deep Learning for Predicting Metastasis on Melanoma WSIs.
\newblock In: 2023 IEEE 20th International Symposium on Biomedical Imaging (ISBI). IEEE; 2023. p. 1--5.

\bibitem{ilse2018attention}
Ilse M, Tomczak J, Welling M.
\newblock Attention-based deep multiple instance learning.
\newblock In: International conference on machine learning. PMLR; 2018. p. 2127--2136.

\bibitem{lu2021data}
Lu MY, Williamson DF, Chen TY, Chen RJ, Barbieri M, Mahmood F.
\newblock Data-efficient and weakly supervised computational pathology on whole-slide images.
\newblock Nature biomedical engineering. 2021;5(6):555--570.

\bibitem{srinidhi2021deep}
Srinidhi CL, Ciga O, Martel AL.
\newblock Deep neural network models for computational histopathology: A survey.
\newblock Medical Image Analysis. 2021;67:101813.

\bibitem{zhang2022dtfd}
Zhang H, Meng Y, Zhao Y, Qiao Y, Yang X, Coupland SE, et~al.
\newblock DTFD-MIL: Double-tier feature distillation multiple instance learning for histopathology whole slide image classification.
\newblock In: Proceedings of the IEEE/CVF Conference on Computer Vision and Pattern Recognition; 2022. p. 18802--18812.

\bibitem{campanella2019clinical}
Campanella G, Hanna MG, Geneslaw L, Miraflor A, Werneck Krauss~Silva V, Busam KJ, et~al.
\newblock Clinical-grade computational pathology using weakly supervised deep learning on whole slide images.
\newblock Nature medicine. 2019;25(8):1301--1309.

\bibitem{zhang2019pathologist}
Zhang Z, Chen P, McGough M, Xing F, Wang C, Bui M, et~al.
\newblock Pathologist-level interpretable whole-slide cancer diagnosis with deep learning.
\newblock Nature Machine Intelligence. 2019;1(5):236--245.

\bibitem{wetteland2021automatic}
Wetteland R, Kvikstad V, Eftest{\o}l T, T{\o}ssebro E, Lillesand M, Janssen EA, et~al.
\newblock Automatic diagnostic tool for predicting cancer grade in bladder cancer patients using deep learning.
\newblock IEEE Access. 2021;9:115813--115825.

\bibitem{yao2020whole}
Yao J, Zhu X, Jonnagaddala J, Hawkins N, Huang J.
\newblock Whole slide images based cancer survival prediction using attention guided deep multiple instance learning networks.
\newblock Medical Image Analysis. 2020;65:101789.

\bibitem{le2023deep}
Le VL, Michot A, Cromb{\'e} A, Ngo C, Honor{\'e} C, Coindre JM, et~al.
\newblock A Deep Attention-Multiple Instance Learning Framework to Predict Survival of Soft-Tissue Sarcoma from Whole Slide Images.
\newblock In: MICCAI Workshop on Cancer Prevention through Early Detection. Springer; 2023. p. 3--16.

\bibitem{liu2024advmil}
Liu P, Ji L, Ye F, Fu B.
\newblock Advmil: Adversarial multiple instance learning for the survival analysis on whole-slide images.
\newblock Medical Image Analysis. 2024;91:103020.

\bibitem{liu2023dsca}
Liu P, Fu B, Ye F, Yang R, Ji L.
\newblock DSCA: A dual-stream network with cross-attention on whole-slide image pyramids for cancer prognosis.
\newblock Expert Systems with Applications. 2023;227:120280.

\bibitem{godson2024immune}
Godson L, Alemi N, Nsengimana J, Cook GP, Clarke EL, Treanor D, et~al.
\newblock Immune subtyping of melanoma whole slide images using multiple instance learning.
\newblock Medical Image Analysis. 2024;93:103097.

\bibitem{chen2020simple}
Chen T, Kornblith S, Norouzi M, Hinton G.
\newblock A simple framework for contrastive learning of visual representations.
\newblock In: International conference on machine learning. PMLR; 2020. p. 1597--1607.

\bibitem{ciga2022self}
Ciga O, Xu T, Martel AL.
\newblock Self supervised contrastive learning for digital histopathology.
\newblock Machine Learning with Applications. 2022;7:100198.

\bibitem{fashi2022self}
Fashi PA, Hemati S, Babaie M, Gonzalez R, Tizhoosh H.
\newblock A self-supervised contrastive learning approach for whole slide image representation in digital pathology.
\newblock Journal of Pathology Informatics. 2022;13:100133.

\bibitem{ke2021contrastive}
Ke J, Shen Y, Liang X, Shen D.
\newblock Contrastive learning based stain normalization across multiple tumor in histopathology.
\newblock In: Medical Image Computing and Computer Assisted Intervention--MICCAI 2021: 24th International Conference, Strasbourg, France, September 27--October 1, 2021, Proceedings, Part VIII 24. Springer; 2021. p. 571--580.

\bibitem{perez2022staincut}
P{\'e}rez JCG, Baguer DO, Maass P.
\newblock Staincut: Stain normalization with contrastive learning.
\newblock Journal of Imaging. 2022;8(7).

\bibitem{li2021dual}
Li B, Li Y, Eliceiri KW.
\newblock Dual-stream multiple instance learning network for whole slide image classification with self-supervised contrastive learning.
\newblock In: Proceedings of the IEEE/CVF conference on computer vision and pattern recognition; 2021. p. 14318--14328.

\bibitem{li2022lesion}
Li J, Zheng Y, Wu K, Shi J, Xie F, Jiang Z.
\newblock Lesion-aware contrastive representation learning for histopathology whole slide images analysis.
\newblock In: International Conference on Medical Image Computing and Computer-Assisted Intervention. Springer; 2022. p. 273--282.

\bibitem{tu2022dual}
Tu C, Zhang Y, Ning Z.
\newblock Dual-Curriculum Contrastive Multi-Instance Learning for Cancer Prognosis Analysis with Whole Slide Images.
\newblock Advances in Neural Information Processing Systems. 2022;35:29484--29497.

\bibitem{wetteland2020multiscale}
Wetteland R, Engan K, Eftest{\o}l T, Kvikstad V, Janssen EA.
\newblock A multiscale approach for whole-slide image segmentation of five tissue classes in urothelial carcinoma slides.
\newblock Technology in Cancer Research \& Treatment. 2020;19:1533033820946787.

\bibitem{fuster2023active}
Fuster S, Khoraminia F, Eftest{\o}l T, Zuiverloon T, Engan K.
\newblock Active Learning Based Domain Adaptation for Tissue Segmentation of Histopathological Images.
\newblock In: 2023 31st European Signal Processing Conference (EUSIPCO). IEEE; 2023. p. 1045--1049.

\bibitem{andersson2014lamina}
Andersson KE, McCloskey KD.
\newblock Lamina propria: the functional center of the bladder?
\newblock Neurourology and urodynamics. 2014;33(1):9--16.

\bibitem{brooks2016positive}
Brooks M, Mo Q, Krasnow R, Ho PL, Lee YC, Xiao J, et~al.
\newblock Positive association of collagen type I with non-muscle invasive bladder cancer progression.
\newblock Oncotarget. 2016;7(50):82609.

\bibitem{wetteland2021parameterized}
Wetteland R, Engan K, Eftesol T.
\newblock Parameterized extraction of tiles in multilevel gigapixel images.
\newblock In: 2021 12th International Symposium on Image and Signal Processing and Analysis (ISPA). IEEE; 2021. p. 78--83.

\bibitem{khosla2020supervised}
Khosla P, Teterwak P, Wang C, Sarna A, Tian Y, Isola P, et~al.
\newblock Supervised contrastive learning.
\newblock Advances in Neural Information Processing Systems. 2020;33:18661--18673.

\bibitem{dietterich1997solving}
Dietterich TG, Lathrop RH, Lozano-P{\'e}rez T.
\newblock Solving the multiple instance problem with axis-parallel rectangles.
\newblock Artificial intelligence. 1997;89(1-2):31--71.

\bibitem{krizhevsky2012imagenet}
Krizhevsky A, Sutskever I, Hinton GE.
\newblock Imagenet classification with deep convolutional neural networks.
\newblock Advances in neural information processing systems. 2012;25.

\bibitem{abraham2019novel}
Abraham N, Khan NM.
\newblock A novel focal tversky loss function with improved attention u-net for lesion segmentation.
\newblock In: 2019 IEEE 16th international symposium on biomedical imaging (ISBI 2019). IEEE; 2019. p. 683--687.

\end{thebibliography}

\end{document}